%% file: main.tex
\documentclass[twoside,11pt]{article}
\usepackage[preprint]{dmlr2e}
\makeatletter
\def\@maketitle{\vbox{\hsize\textwidth
 \linewidth\hsize \vskip \beforetitskip
 {\begin{center} \Large\bf \@title \par \end{center}} \vskip \aftertitskip
 {\def\and{\unskip\enspace{\rm and}\enspace}%
  \def\addr{\small\it}%
  \def\email{\hfill\small\sc}%
  \def\name{\normalsize\bf}%
  \def\AND{\@endauthor\rm\hss \vskip \interauthorskip \@startauthor}
  \@startauthor \@author \@endauthor}
  \vskip \aftermaketitskip
}}
\makeatother

\usepackage{amsmath}
\usepackage{booktabs}
\usepackage{multirow}
\usepackage{array}
\usepackage{xcolor}
\usepackage{url}
\usepackage{siunitx}
\usepackage{tikz}
\usetikzlibrary{arrows.meta,calc,positioning}
\usepackage{pgf}
\usepackage{pgfplots}
\pgfplotsset{compat=1.18}
\usepackage{listings}
\usepackage{adjustbox}
\usepackage{inconsolata}
\usepackage{microtype}
\usepackage{lastpage}
\hypersetup{
    pdftitle={Generating Schema-Aligned Synthetic Dialogues and Measuring Their Diversity},
}

\dmlrheading{}{2026}{1-\pageref{LastPage}}{}{}{}{Konrad Stolte and Toni Fetzer}
\ShortHeadings{Schema-Aligned Synthetic Dialogues and Their Diversity}{Stolte and Fetzer}
\firstpageno{1}

\begin{document}

\title{Generating Schema-Aligned Synthetic Dialogues and Measuring Their Diversity}

\author{\name Konrad Stolte \email konrad.stolte@cronn.de \\
      \addr cronn GmbH, Bonn, Germany
      \AND
      \name Toni Fetzer \email toni.fetzer@cronn.de \\
      \addr cronn GmbH, Bonn, Germany}

\maketitle

\begin{abstract}%
Training or evaluating a model that extracts structured information from conversations requires pairs of transcripts and their corresponding records.
Creating these pairs by hand is expensive, and in sensitive domains, they are also difficult to obtain.
Instead, we reverse the process and generate them in the opposite order by sampling a schema-valid record first, then writing a dialogue from it, and optionally synthesizing speech from the dialogue.
This way, we already have the correct labels and do not need manual annotation.
Simply prompting a language model for such dialogues tends to yield very similar conversations with little variation, so we add a sampled variation configuration, personas, and a topic plan that determines the order in which information appears.
We measure the variation using text embeddings, and find that the embedding variance spreads over about two and a half times as many independent directions with our full configuration compared to without (participation ratio \num{23.3} against \num{9.2}).
Two medical dialogue corpora show values in a similar range, although sample size, text length, and language also affect the results.
We also show the beneficial effect of our method in a downstream test, where an 8B extraction model reaches \SI{64.8}{\percent} field accuracy when fine-tuned on dialogues from our method, against \SI{57.5}{\percent} from naive generation.
We release both the generation library and a generated dataset of German clinical dialogues with aligned records.
{\renewcommand{\thefootnote}{\fnsymbol{footnote}}%
\footnote[1]{The library and the dataset are available at \url{https://github.com/cronn/conversim}.}}
\end{abstract}


\section{Introduction}
Many tasks involve mapping a spoken or written conversation to a document with a fixed structure of sections and fields.
Examples of this include medical appointments, emergency calls or interviews in general.
Training or evaluating such a model requires conversations paired with the filled-in record that they should produce.
However, these pairs are hard to collect, annotating them manually is slow, and in sensitive domains the conversations cannot be shared or sent to a third-party service either.
Using an existing model for annotation would just result in distillation of that model, and especially in a specific and difficult domain, finding a sufficiently good teacher model also becomes an issue.

Our approach is to build the data in the opposite order.
We first generate a filled-in record that follows our target schema, then create dialogue based on that record, and optionally synthesize speech from that dialogue (Figure~\ref{fig:synthetic-pipeline}).
No real consultation is recorded or annotated, and no real patient's data enters the dataset, which is very beneficial for privacy and data protection.
This is not enough to create useful data, though, because simply asking a language model to create a realistic conversation from a record will only produce small variations of one conversation.
Section~\ref{sec:pipeline} describes the mechanisms we add against this, and they are also generally applicable to other schemas.

\input{figures/synthetic-pipeline}

The next step is to find out whether these methods actually improve diversity.
Diversity is the most important property here, because a generator that writes the same or very similar conversation 452 times would give us a large dataset, but not a useful one for training.
Having many possible generation settings does not automatically mean that the resulting conversations are diverse and provides only an upper bound, so Section~\ref{sec:diversity} measures the variation that arrives in the text.

To do this, we embed each conversation and analyze how these embeddings are distributed.
We look at the geometry of the resulting point cloud, and compare the full configuration against a narrower generation and against two corpora of medical dialogue.
Since these measurements can be affected by the number of conversations, or their length, we also examine and report these effects.

For our case study, we use German medical conversations about sexually transmitted infections (STIs) and HIV.
A medical professional talks to a patient about sexual history, possible exposure, symptoms, etc., and the result is filled into an anamnesis form.
These conversations run for several minutes, include specialized medical vocabulary, and deal with sensitive personal topics.
This makes real conversations hard to obtain and is therefore a real use case for our method.
In this work, we mainly focus on generating and evaluating the data itself, and we include a small experiment to test how useful the data is for training a model (Section~\ref{sec:utility}).

Our contributions are as follows.
We describe a pipeline that generates conversations from structured records and adds methods to make these conversations more varied.
We measure the diversity of the generated data in embedding space, and we document how the statistics depend on sample size, text length and language.
We compare our full generation setup with a less varied version and with two existing datasets, MedDialog and PriMock57, while also discussing the limits of these comparisons.
We also test the impact of our generation setup on model performance in an experiment using differently varied dialogues, and the results support the advantage of our method.
Finally, we release the generation pipeline as the open-source library conversim, including our STI/HIV configuration, as well as the generated dataset of 452 German clinical dialogues, with aligned records for the 401 dialogues of the full configuration, and 28 dialogues with hand-corrected transcripts and records.

\section{Related Work}
The idea of generating annotations before writing the actual conversation is not a new one.
\citet{shah2018m2m} use dialogue self-play to produce an outline of semantic frames, which is then rewritten by crowdworkers into natural utterances.
The Schema-Guided Dialogue dataset uses the same idea at a larger scale, with over \num{16000} conversations across 16 domains, combining a dialogue simulator with paid crowdworkers~\citep{rastogi2020sgd}.
In this case the schema describes how a service works, including its intents, slots, and their descriptions.
This allows models to work with services they have not seen before.
But both of these datasets build the dialogue outline from rules and templates, and only its wording comes from crowdworkers.
Our approach differs, because our schema describes the structured record behind the entire conversation instead of a service interface, and we use a language model to generate the dialogue itself.

Looking at the clinical domain, similar approaches have also been used there, although they usually start with free-text notes rather than structured records.
Concretely, MedSynth writes a note first and the dialogue from it, also releasing over \num{10000} dialogue-note pairs~\citep{medsynth2025}.
SynDial starts from clinical notes that already exist and generates the dialogue from them~\citep{syndial2024das}.
But in both cases, the basis of the dialogue is a written note and not a typed record.

Regarding diversity, MedSynth also tries to control it by rejecting a new scenario unless at least four of its thirteen scenario variables differ from already selected ones~\citep{medsynth2025}.
We take a different approach, because instead of filtering scenarios afterwards, we define different sources of variation in advance and sample them for each conversation.
We then measure how much variation actually appears in the generated text via empirical means (Section~\ref{sec:diversity}).
A typed record gives us more precise control over the data, but we need to consider how to handle empty fields and actual conversational flow.
We go more into detail in Sections~\ref{sec:schema} and~\ref{sec:variation} on this.

For speech generation, \citet{synthdopaco2026} have developed synthetic doctor-patient datasets using an audio generation approach similar to ours.\footnote{Our work took place in parallel to \citet{synthdopaco2026}, and the two were developed independently of each other.}
However, their pipeline starts by generating the dialogue and then creates clinical notes from it, while we start with the structured record.
Synthetic data is in any case a well-established method to train modern speech models~\citep{canary2024puvvadaa}.

Finally, there currently exists no public dataset which contains long German dialogue audio, with transcripts and an aligned coded record.
German speech datasets are mostly neither medical nor conversational~\citep{commonvoice2020ardila,asrbundestag2023wirth}, and even when looking at other languages~\citep{meddialog2020he} or domains~\citep{amimeetingcorpus2006carletta}, medical speech data tends to consist of only short utterances~\citep{MultiMed-ST2025le-duc}, lacks audio, or is limited in context.

\section{Generation Pipeline}\label{sec:pipeline}
The pipeline has one initial design stage and three generation stages and can be seen in Figure~\ref{fig:synthetic-pipeline}.
First, the schema is designed by hand.
For each new datapoint, we then generate a record based on the schema, use the record to write a dialogue, and optionally synthesizing speech from the dialogue.
The pipeline itself is not tied to any specific domain and users can define their own schemas, participant roles, personas, variation settings, and topic categories.
The dataset presented in Section~\ref{sec:corpus} is therefore just one example of how the pipeline can be used.

We use different models for the individual generation stages.
For sampling records, we chose a fast model that could reliably produce structured output (Gemini~2.5~Flash at the time), and we additionally validate its output against the schema before accepting it.
For the dialogue we used a different model that writes better conversational text (GPT-5-mini at the time).
Splitting the stages this way allows us to easily isolate failures and keep exact control of the flow of information through the pipeline.

\subsection{Schema Design}\label{sec:schema}
The schema is defined as a typed Pydantic model~\citep{pydantic}, with each field having a descriptive name and a short explanation of what information it should contain.
Handing this schema to the generator constrains it to only produce valid records, considering both the data types and the semantic constraints.
The same schema can later be handed to a model that extracts the record from conversations and further used to validate and score that output.

The granularity of the schema needs to be considered carefully and one needs to find the right balance.
Our first schema was coarse, with high-level fields and a lot of free text, which gives too much freedom to the generator and makes the extraction unscorable, while a second version with confidence values, more fine-grained typing and evidence metadata was too complex for our local models and used too many tokens.
Our final schema is a compromise between these two versions, using structured, coded fields but leaving out detailed metadata.
This gives us enough structure to guide the models and evaluate their output, while keeping the schema flexible and manageable.
Figure~\ref{fig:schema-example} shows an example of a group of fields and a generated record.

\input{figures/schema-example}

Since we use a typed schema, we also need to consider how to represent the absence of data and with that the fact that not every conversation needs to cover every subject.
For this, a field can be marked as \texttt{not\_discussed} if the topic never comes up in the conversation, \texttt{unknown} if the topic comes up but no clear answer is given, and \texttt{not\_applicable} if the field does not make sense in the given context.
We also explicitly tell the dialogue generator not to mention any fields marked as \texttt{not\_discussed}.
This helps create more realistic conversations in which not every possible topic has to be covered.
The corresponding prompt instruction is shown in the top-right block of Figure~\ref{fig:prompt-mechanisms}.

\input{figures/prompt-mechanisms}

\subsection{Controlled Variation}\label{sec:variation}
If we simply asked a large language model (LLM) for a realistic conversation repeatedly, the results would often be quite similar.
The randomness of the model alone is not enough to create a large and diverse dataset, and we therefore introduce several ways to control how the conversations differ from one another.

For this, we include a \emph{structured variation configuration} in the prompt (see Figure~\ref{fig:synthetic-pipeline}).
It specifies additional properties that change the flow of the conversation, and we choose seven conversation types, seven conversation styles, nine coarse patient profiles, three health-literacy levels, six discourse structures, five settings and three dialects (applied to half of the conversations), together with flags for, e.g., opening small talk (\SI{20}{\percent}), operator-initiated explanation (\SI{30}{\percent}), or participant-initiated questions (\SI{40}{\percent}).

In addition to that, we randomly draw two personas from a pool of five clinicians and eight patients.
These personas describe how someone should talk and, for example, how forthcoming they are, so the conversations become more dynamic and realistic.

The selected variation settings are included in both the structured record and dialogue generation.
This helps the record generator to avoid inconsistencies and sample a randomized high-level outline for the dialogue, and the dialogue generator to gain additional context for the dialogue.

In order to further extend the variation, we also add a topic plan, since the record does not specify any ordering.
For our medical domain, we define 15 topic categories, each one linked to specific fields in the schema and including a description, a priority, and rules for when it should be discussed.
A planner then orders and prioritizes the categories (categories without content are dropped) and also includes them in the prompt to ensure a natural conversational flow, as shown in the bottom right block of Figure~\ref{fig:prompt-mechanisms}.
We finally validate the field paths against the schema and make sure to only allow valid samples to pass through the pipeline.

\subsection{Speech Synthesis}\label{sec:speech}
The speech synthesis is done in a separate pass over the final dialogues, since it requires more computational resources.
We synthesize one segment per utterance, which keeps our speaker assignment clear and allows us to further tweak the audio during assembly.
The conversation is assembled with stereo placement, randomized pauses between turns (that may also overlap slightly), small gain variations, room tone and probabilistic listener acknowledgements.
We use the open-weight system Qwen3TTS with voice cloning~\citep{qwen3tts2026hu}, which has shown promising results.

Because we generate each utterance separately, we can also create two audio versions of the same conversation by swapping the speakers' voices.
This does not increase the amount of unique dialogue text, but it doubles the available data for speech-specific training.
Another benefit is that we know the exact start and end of each speaker turn, which makes the data potentially useful for speaker diarization tasks as well.

\subsection{Determinism and Cost}
To make our pipeline reproducible, we calculate a separate random seed for each sample based on a root seed and the sample index.
This means that the same sample can be generated independently of the order in which samples are processed and if a generation run fails, we can continue from where it stopped without having to generate the successful samples again.

Generating the dialogue text is also inexpensive.
Based on the actual prompt and output lengths in our dataset and current model pricing, generating a conversation costs less than one cent.

\section{Case Study Dataset}\label{sec:corpus}
Our schema for the case study of German STI/HIV anamnesis covers sexual history, exposure risk, symptoms, prior testing, prevention and the resulting test recommendations.
We based it on different publicly available anamnesis forms and guidelines and turned it into a Pydantic model, as described in Section~\ref{sec:schema}.
We also discussed the schema with a counselor at the Gesundheitsamt Frankfurt (public health office) who conducts these consultations himself.
He considered both the schema and the conversations as realistic for the setting.

Our dataset, StiAna, contains 452 dialogues with aligned records, with an average of 40 turns per dialogue and \num{304419} words in total.
It consists of two parts, which we analyze separately in Section~\ref{sec:diversity}.
StiSimple is a set of 51 dialogues that we generated with an earlier version of the schema, before introducing our variation configuration and topic plan, and StiCloned is a set of 401 dialogues generated with the mechanisms of Section~\ref{sec:variation}.
A further held-out split, StiReal, contains 28 generated scripts that were acted out by human speakers, whose transcripts we corrected by hand against what was actually said.
We use this dataset as an additional reference in our embedding analysis in Section~\ref{sec:diversity} and to evaluate model performance in Section~\ref{sec:utility}.

We publish our generation pipeline and its configuration as the open-source library \emph{conversim}.
We also release the generated dialogues, the structured records for StiCloned, and the manually corrected transcripts and records for StiReal.
The records from StiSimple are not included because they use an older version of the schema.
We do not release the recorded audio, because the speakers did not consent to distribution.
The same goes for the voice-cloned audio, since it is almost identical to the original voice.
However, the synthetic speech can be recreated using one's own voice and requires only the resources to run Qwen3TTS~\citep{qwen3tts2026hu}.

\section{Diversity of the Generated Data}\label{sec:diversity}
We need to both look at how much variation is theoretically possible with our design as well as how much variation we actually see in the generated conversations in order to check that our pipeline produces a diverse dataset.

\subsection{Nominal Variation}
If we just look at the categorical schema fields independently, we get roughly \num{8.56e19} combinations.
Aggregating them into clinically appropriate profiles still leaves us with around \num{1.21e12}, where we group ten combinations of sex, gender identity and anatomy crossed with three age brackets for the patient, and partner count, practices, condom use and partner status for the exposure, and treat symptoms, prior history, prevention, substance use, risk and testing in the same way.
However, these numbers only describe how many different combinations of facts are possible, and changing individual facts does not necessarily make the resulting conversation feel different.
We therefore also look at how much variation our topic planner introduces.
It orders a set of up to 15 topic categories (about ten per conversation), and if we sample \num{50000} orderings from it, we get about \num{3800} meaningfully distinct ones.
We consider two patterns different if they contain a different number of topics or if the order of their first five topics differs.
We focus on these first five topics because they largely determine how the conversation develops.
This gives us a lower bound on the number of different conversation structures our planner can produce and shows that topic ordering adds another source of variation.

\subsection{Realized Variation in Embedding Space}
But since theoretical analysis will only go so far, and we therefore also analyzed the actual variation in the generated data.
For this, we embed all 452 conversations with \texttt{intfloat/\allowbreak multilingual-e5-large}~\citep{multilinguale5}.
We embed every sentence separately and average the sentence embeddings of a conversation, which gives us one point per conversation in \num{1024} dimensions.
We then run a principal component analysis (PCA) over these points to study how these embeddings are distributed.
Our results show that much of the variation can be described using fewer than the full \num{1024} dimensions.
We need 137 principal components to explain \SI{90}{\percent} of the variance and 200 components to explain \SI{95}{\percent}.
These correspond to \SI{30}{\percent} and \SI{44}{\percent} of the 451 directions that are possible with 452 centered data points.
The variance is not concentrated in a handful of directions, though, because reaching \SI{99}{\percent} still takes 323 components (\SI{72}{\percent} of the 451) and the participation ratio is \num{31.1}.
The participation ratio $(\sum_i \lambda_i)^2 / \sum_i \lambda_i^2$ over the PCA eigenvalues $\lambda_i$ counts how many equally strong directions a given spread of variance corresponds to.
All variance in one direction results in a 1, an even spread over $k$ directions gives $k$.
Across the \num{1024} coordinates of the embedding space, the least varying coordinate still has about a third of the average variance per coordinate.
We also measure how far each conversation lies from the center of all embeddings.
The average Euclidean distance is \num{0.057}, with a coefficient of variation of \num{0.22}.
A few conversations are noticeably further away, with the largest distance reaching \num{0.150}.

An important property of PCA is that these component counts grow with the sample size, because a PCA over $n$ points has at most $n-1$ directions to work with.
StiCloned alone needs 34 components for \SI{90}{\percent} of its variance at $n=51$, but 139 at $n=401$.
Counts are therefore only comparable at the same $n$, and all comparisons below are made at a matched $n$ unless stated otherwise.

As a straightforward comparison, we compare StiSimple ($n=51$) against StiCloned.
StiSimple was generated without the variation configuration and the topic planner.
Both datasets contain long German medical conversations, and we embedded them together using the same model and settings.
This makes the comparison more consistent, since we do not have to account for language differences and the difference in conversation length is relatively small.
StiSimple conversations are longer on average, with 846 against 652 words.
As we show in Section~\ref{sec:human-comparison}, conversation length can affect the participation ratio, but even reducing the number of sentences per conversation from 71 to 8 only changes the participation ratio by a factor of \num{1.4}.
The difference in average length between StiSimple and StiCloned is much smaller, with a ratio of about \num{1.3}.
It is therefore unlikely that conversation length alone can explain the difference in diversity we observe below.

\input{figures/diversity-figures}

Figure~\ref{fig:variance-curves} shows the result at a matched $n=51$, together with the two human corpora that we discuss further in Section~\ref{sec:human-comparison}.
StiSimple has the steepest curve of all four datasets and explains \SI{90}{\percent} of its variance in only 28 components, compared to 34 for StiCloned.
The same difference appears in the participation ratio, where StiSimple reaches \num{9.2} and StiCloned \num{23.3}.
This means that the variation in StiCloned is spread across roughly two and a half times as many equally important directions.

To check how stable this result is, we randomly selected 51 conversations from the 401 StiCloned dialogues 200 times.
The resulting interval of the participation ratio spans \numrange{20.3}{26.3}, so the narrow value of \num{9.2} lies far below it.
However, StiSimple contains only 51 conversations, so we cannot calculate a similar interval for it at this sample size.
Regarding absolute dispersion, the narrow subset shows slightly larger Euclidean pairwise and radial distances (\num{0.089} against \num{0.076}, and \num{0.062} against \num{0.053}).
StiSimple is only a single draw of 51 conversations, so it has no interval of its own at this $n$.

This suggests that the variation space spreads the variance over more independent directions, instead of only widening the few directions the narrow run already had.
When looking at the full dataset instead of a subsample, StiCloned has a participation ratio of \num{41.5} ($n=401$), and adding in the 51 conversations from StiSimple lowers it to \num{31.1} overall.
The drop is mainly due to the difference between the two groups, which explains \SI{9}{\percent} of the total variance and is almost entirely aligned with the first principal component.
This is also why the full dataset needs fewer components for \SI{90}{\percent} of its variance than StiCloned alone (137 against 139), even though it has more points.
Figure~\ref{fig:tsne-overlay} shows the clustering of our datasets directly with a t-SNE plot~\citep{tsne2008vandermaaten} for illustration purposes.
In this projection, StiCloned forms one broad cloud with StiReal inside it, while StiSimple appears as a separate group.

\subsection{Comparison with Human Dialogue}\label{sec:human-comparison}
To better understand how diverse our generated conversations are, we also provide a comparison with real human dialogue and look at two reference corpora.
The first is MedDialog~\citep{meddialog2020he}, which consists of 603 real doctor-patient exchanges.
These conversations use genuine human language but are much shorter than ours, at about 8 sentences per exchange against our 71.
The second is PriMock57~\citep{primock57}, which consists of 57 primary-care consultations between real clinicians and trained role-players.
They were recorded and transcribed by hand, and at 1507 words on average these are about 2.3 times as long as ours (652 words for StiCloned).
Both reference datasets are English, while ours is German, and both cover different medical topics.

Our embedding model supports multiple languages, so we can use it for both English and German conversations.
However, conversation length also affects the embeddings, and since we average the embeddings of individual sentences, longer conversations tend to have smaller distances between their final embeddings.
More specifically (assuming independent sentence embeddings), the average distance from the center is expected to decrease approximately with $1/\sqrt{N}$, where $N$ is the number of sentences, which would be a factor of 3 between MedDialog and ours.
Re-embedding our conversations from only 8 sentences each raises the mean radial distance from \num{0.053} to \num{0.151}, a factor of \num{2.8}, close to this prediction, and moves it towards MedDialog (\num{0.201}).
The distances are therefore only comparable between datasets with very similar conversation lengths, so we instead focus on how the variance is distributed across different directions.
However, conversation length also influences these measurements, because when we reduce StiCloned to 8 sentences per conversation, using all 401 conversations, its participation ratio increases from \num{41.5} to \num{59.3}, and the number of components needed to explain \SI{90}{\percent} of the variance increases from 139 to 148.

At a matched $n=51$ (Figure~\ref{fig:variance-curves}), StiCloned and MedDialog are almost the same in their variance curves.
They need 34 and 33 components for \SI{90}{\percent} of the variance, with participation ratios of \num{23.3} (interval \numrange{20.3}{26.3}) and \num{24.3} (interval \numrange{21.2}{27.1}), so the intervals overlap almost completely.
PriMock57 is more concentrated than both, meaning that its variance sits in fewer directions, with 29 components and a participation ratio of \num{18.3} (interval \numrange{17.6}{19.2}, which is narrow because it draws 51 of only 57 conversations).
StiSimple has its variance in the fewest directions of all of them.
However, we need to be careful when interpreting these comparisons because of the differences in conversation length.
For example, at $n=401$ and full length, MedDialog has the higher participation ratio (\num{53.0} against our \num{41.5}), but when our conversations are re-embedded from 8 sentences each to match MedDialog's length, ours is the higher one (\num{59.3} against \num{53.0}).
This means that which dataset appears more diverse depends on how we account for conversation length and we therefore cannot use these numbers to reliably rank the datasets by diversity.

Overall, our results show that, at the same sample size, StiCloned has a participation ratio in a similar range to the two human dialogue datasets.
We therefore find no evidence that our generated conversations are much less varied in embedding space than the human medical dialogues we compare them with.
In contrast, StiSimple, which was generated without our additional variation methods, shows clearly less variation across independent directions.

\section{Does the Variation Matter Downstream?}\label{sec:utility}
Since embedding measurements only give us an indirect indication of diversity, we also run a small experiment to verify our findings and see how the additional variation helps when training a model.
For this, we fine-tune a local extraction model to produce the record from the dialogue text, and compare two training sets of the same size, all 51 StiSimple dialogues and 51 dialogues drawn at random from StiCloned.
We use Apertus~8B Instruct~\citep{apertus2025}, an open multilingual model whose weights and training data are both public.
We evaluate on the 28 StiReal dialogues, using the manually corrected transcripts as input.
The metric is the field accuracy over the schema, which is the share of correctly extracted fields, with partial credit for list fields and with a required schema-valid output.

Training uses 4-bit QLoRA~\citep{qlora2023dettmers} (rank 16, $\alpha=16$, all attention and feed-forward projections, learning rate $2\times10^{-4}$, effective batch size 8) for three epochs, with one run per training set.
We merge the adapter, quantize the model to \texttt{q4\_k\_m} and decode with llama.cpp at temperature 0.2 under a JSON-schema grammar.
The untrained model is quantized and evaluated in the same way.

\begin{table}[t]
  \centering
  \small
  \begin{tabular}{lc}
    \toprule
    Model & Field accuracy (\%) \\
    \midrule
    Apertus~8B, untrained & \num{52.5} \\
    \quad fine-tune on 51 StiSimple dialogues & \num{57.5} \\
    \quad fine-tune on 51 StiCloned dialogues & \num{64.8} \\
    \bottomrule
  \end{tabular}
  \caption{Field accuracy of Apertus~8B Instruct on 28 StiReal dialogues, before and after fine-tuning on 51 dialogues from either part of the dataset.}
  \label{tab:utility}
\end{table}

Table~\ref{tab:utility} shows that both training sets improve the model, but training on StiCloned leads to a much larger improvement.
Fine-tuning on StiSimple increases field accuracy by \num{5.0} percentage points compared to the untrained model (bootstrap interval over the dialogues \numrange{1.4}{8.5}), and fine-tuning on StiCloned improves it by \num{12.3} points (bootstrap interval \numrange{8.3}{16.3}).
The difference between the two training sets is \num{7.3} points (\numrange{4.5}{10.4}), and the StiCloned model is better on 22 of the 28 dialogues.
These results show that the additional variation in StiCloned is useful beyond the embedding measurements and even with the same number of training dialogues, the full variation configuration produces better training data for this extraction task.

\section{Discussion and Conclusion}
Our results suggest that if the output of a task is a typed record, the record should be generated first.
This way, each conversation already has a matching ground-truth record, so there is no need to annotate it afterwards.
It also gives us more control over how the conversations are generated and allows us to introduce different sources of variation.

One of the most important parts of this approach is designing the schema carefully.
Since the schema is used throughout the generation process and later for evaluation, its design has a large influence on the results.
It needs to handle optional fields/missing information and still keep a good balance between detailed data types and flexibility.
It is also important to consider from the beginning how the extracted records will eventually be evaluated.

Our results and measurements also show that the additional variation methods lead to more diverse conversations.
At a matched sample size, the participation ratio of the embedded conversations is about two and a half times higher with the variation mechanisms than without them.
A caveat is that having many possible combinations of settings does not automatically result in the same realized variation, and we therefore need to measure the actual variation in the generated text.
These measurements need to be interpreted carefully, because they depend on the sample size and the number of sentences per conversation.
Specifically, we have observed that the length affects not only how far apart the embeddings are, but also how their variation is distributed across different directions.

Our extraction experiment supports these findings.
At 51 dialogues, the full configuration gives about 7 points higher field accuracy than the run without it, and more than twice the gain over the untrained model.
Although this is only a small experiment, it suggests that the additional variation can make the generated data more useful for training.

Finally, an important and practical advantage of our approach is that none of the pipeline is specific to our domain.
We therefore release the generator together with the dataset, so that the same approach may be used to create data for other domains and structured reporting tasks.

\section{Limitations}\label{sec:limitations}
Our analysis covers one schema, one language, and one embedding model, with datasets of up to 452 conversations.
Future work should therefore look at whether our results also hold for other languages and larger datasets.

The comparison between StiSimple and StiCloned also has some limitations.
StiSimple uses an older version of the schema, and its conversations are longer.
This means we cannot be sure that the differences we found are only caused by the variation methods.

The extraction experiment has some limitations as well.
Our evaluation set contains only 28 conversations because creating a high-quality, manually checked dataset takes a lot of time and effort.
Also, the evaluation scripts were generated using the full configuration, which might give the model trained on StiCloned an advantage.
We also only trained each model once per dataset.

Future work should also include different embedding models, to analyze a potential impact from this angle, further datasets, which could be created by translation or with our generation approach, and more downstream models with multiple training runs.

\impact{
We did not use any real patient data in this work.
Instead, we created fictional patient records based on common medical history forms and generated fictional dialogues.
The voice recordings were made by human speakers who agreed to be recorded, but did not agree to share their recordings publicly.
For this reason, we do not release any audio data, including the voice-cloned recordings.

When we created the dataset, we wanted to include a wide range of patient profiles.
These differ, for example, in sexual orientation, gender identity, sex work, personal background, and age, because we believe that this diversity is important for creating a high-quality dataset that teaches a model about many different types of patients.
The dataset is intended to be used for fine-tuning LLMs for this task.

Using synthetic data also helps to reduce any privacy concerns, since no real patient information is used to train the model.
However, the results still need to be checked carefully before the system can be used in practice.
Even if the output looks correct, some fields may contain information that seems reasonable but is actually wrong.

One limitation of synthetic data is that it can contain the same biases as the model that generates it.
We tried to mitigate this by using the different patient personas and variation settings, however, the language and medical content are still influenced by the generating model.
A model trained on this data may therefore learn some of the same biases and weaknesses.

Another risk would be if synthetic dialogues would completely replace validation on real medical consultations, because a system should also be evaluated using real data before it is used in practice.

Finally, voice cloning also comes with ethical risks, as it can be used to copy someone's voice without their permission, but in our work, we only cloned the voices of speakers with their explicit consent.
We also decided not to share the generated audio recordings publicly.
}

\acks{
This work was carried out as part of the first author's master's thesis at the University of Bonn in cooperation with cronn GmbH.
cronn GmbH provided access to the commercial language model APIs and the GPUs used for training.
Konrad Stolte was employed at cronn GmbH as a student employee during this work.
Toni Fetzer is employed at cronn GmbH and supervised the thesis there.
The authors declare no other competing interests.
We thank the staff member at the Gesundheitsamt Frankfurt who reviewed the schema and the dialogues, and the speakers who acted and recorded the StiReal dialogues.
}

\vskip 0.2in

\bibliography{custom}

\end{document}

%% file: figures/synthetic-pipeline.tex
\begin{figure*}[t]
	\centering
	\resizebox{0.92\linewidth}{!}{%
		
		\begin{tikzpicture}[
			node distance=1.2cm and 0.9cm,
			>=Latex,
			font=\small,
			prepbox/.style={
				draw,
				rounded corners,
				thick,
				align=center,
				minimum width=3.3cm,
				minimum height=2.3cm,
				fill=purple!6
			},
			stepbox/.style={
				draw,
				rounded corners,
				thick,
				align=center,
				minimum width=3.5cm,
				minimum height=2.6cm,
				fill=blue!7
			},
			audiobox/.style={
				draw,
				rounded corners,
				thick,
				align=center,
				minimum width=3.3cm,
				minimum height=2.6cm,
				fill=blue!4
			},
			cfgbox/.style={
				draw,
				rounded corners,
				thick,
				dashed,
				align=left,
				minimum width=8.8cm,
				minimum height=2.0cm,
				fill=orange!8
			},
			outputbox/.style={
				draw,
				rounded corners,
				thick,
				align=center,
				minimum width=2.1cm,
				minimum height=1.0cm,
				fill=green!6
			},
			mainarrow/.style={thick, ->},
			influarrow/.style={thick, dashed, ->}
			]
			
			\node[prepbox] (schema) {
				\begin{tabular}{@{}l@{}}
					\multicolumn{1}{c}{\textbf{Schema Design}}\\[1mm]
					\small Define STI/HIV schema
				\end{tabular}
			};
			
			\node[stepbox, right=of schema] (step1) {
				\begin{tabular}{@{}l@{}}
					\multicolumn{1}{c}{\textbf{Ground Truth Generation}}\\[1mm]
					\small Sample schema instance\\
					\small Generate fictitious patient JSON
				\end{tabular}
			};
			
			\node[stepbox, right=of step1] (step2) {
				\begin{tabular}{@{}l@{}}
					\multicolumn{1}{c}{\textbf{Dialogue Script Generation}}\\[1mm]
					\small Generate dialogue based on\\
					\small structured ground truth
				\end{tabular}
			};
			
			\node[audiobox, right=of step2] (step3) {
				\begin{tabular}{@{}l@{}}
					\multicolumn{1}{c}{\textbf{Audio Synthesis}}\\[1mm]
					\small Utterance-level TTS\\
					\small Post-processing
				\end{tabular}
			};
			
			\node[anchor=north west, font=\scriptsize\bfseries] at (schema.north west) {Step 0};
			\node[anchor=north west, font=\scriptsize\bfseries] at (step1.north west) {Step 1};
			\node[anchor=north west, font=\scriptsize\bfseries] at (step2.north west) {Step 2};
			\node[anchor=north west, font=\scriptsize\bfseries] at (step3.north west) {Step 3};
			
			\node[outputbox, above=0.55cm of step1] (o1) {\textbf{Record}\\(JSON)};
			\node[outputbox, above=0.55cm of step2] (o2) {\textbf{Dialogue}\\(script)};
			\node[outputbox, above=0.55cm of step3] (o3) {\textbf{Audio}\\(synthetic)};
			
			\coordinate (mid12) at ($(step1)!0.5!(step2)$);
			\node[cfgbox, below=2.5cm of mid12] (cfg) {
				\begin{tabular}{@{}l@{}}
					\multicolumn{1}{c}{\textbf{Variation Configuration}}\\[1mm]
					\small Randomly sampled conversation type/style/register,\\
					\small health literacy, discourse structure, setting, etc.\\
				\end{tabular}			
			};
			
			\draw[mainarrow] (schema) -- (step1);
			\draw[mainarrow] (step1) -- (step2);
			\draw[mainarrow] (step2) -- (step3);
			
			\draw[mainarrow] (step1) -- (o1);
			\draw[mainarrow] (step2) -- (o2);
			\draw[mainarrow] (step3) -- (o3);
			
			\draw[influarrow] (cfg.north) to[out=90,in=-90] (step1.south);
			\draw[influarrow] (cfg.north) to[out=90,in=-90] (step2.south);
			
		\end{tikzpicture}
		
	}
	\caption{Overview of the synthetic data generation pipeline, which works in the reverse order of how the data will later be used. Pink shows the initial manual schema design, blue the generation steps for each sample (structured record, dialogue, and optional audio), and green the resulting outputs. The dashed orange box shows the variation settings, which steer both record and dialogue generation.}
	\label{fig:synthetic-pipeline}
\end{figure*}

%% file: figures/schema-example.tex
\definecolor{vscodeBlue}{RGB}{86,156,214}     %
\definecolor{vscodeGreen}{RGB}{106,153,85}    %
\definecolor{vscodeGray}{RGB}{128,128,128}    %
\definecolor{vscodeBg}{RGB}{245,245,245}      %
\definecolor{vscodeType}{RGB}{77,201,164}
\definecolor{vscodeFunction}{RGB}{195,168,248}

\lstdefinestyle{vscode-python}{
	language=Python,
	basicstyle=\ttfamily\scriptsize,
	frame=single,
	rulecolor=\color{black},
	breaklines=true,
	breakatwhitespace=true,
	showstringspaces=false,
	tabsize=4,
	numbers=none,
	keywordstyle=\color{vscodeBlue}\bfseries,
	stringstyle=\color{vscodeGreen},
	commentstyle=\color{vscodeGray},
	backgroundcolor=\color{vscodeBg},
	aboveskip=0pt,
	belowskip=0pt,
	emph={BaseModel,Enum,CondomUsePattern,str,int,float,bool,list},
	emphstyle=\color{vscodeType},
	emph={[2]Field},
	emphstyle={[2]\color{vscodeFunction}},
	literate=
	{ä}{{\"a}}1
	{ö}{{\"o}}1
	{ü}{{\"u}}1
	{Ä}{{\"A}}1
	{Ö}{{\"O}}1
	{Ü}{{\"U}}1
	{ß}{{\ss}}1
}

\begin{figure}[t]
\centering
{\ttfamily\scriptsize\color{vscodeGray}schema.py}\par
\vspace{2.7pt}
\begin{lstlisting}[style=vscode-python]
class CondomUsePattern(str, Enum):
    always = "always"
    mostly = "mostly"
    sometimes = "sometimes"
    never = "never"
    not_discussed = "not_discussed"

class PartnerAndExposure(BaseModel):
    condom_use_vaginal: CondomUsePattern = Field(
        default=CondomUsePattern.not_discussed,
        description="Kondomnutzung bei vaginalem Verkehr.")
    # ...
\end{lstlisting}

\vspace{6pt}
{\ttfamily\scriptsize\color{vscodeGray}ground\_truth.json}\par
\vspace{2.7pt}
\begin{lstlisting}[style=vscode-python]
"partners_last_3_months": 2,
"practices_last_3_months": ["vaginal", "oral_insertive"],
"condom_use_vaginal": "sometimes",
"condom_use_anal": "not_discussed",
"partner_hiv_status_known": "not_discussed"
\end{lstlisting}
\caption{The schema definition (top) and an example of a generated record (bottom). Using an enum instead of free text makes the answer scorable. The value \texttt{not\_discussed} indicates that a topic was never mentioned, and the dialogue generator is instructed to leave it out.}
\label{fig:schema-example}
\end{figure}

%% file: figures/prompt-mechanisms.tex
\definecolor{promptFrame}{RGB}{120,120,120}
\definecolor{promptBg}{RGB}{247,247,247}
\definecolor{promptHead}{RGB}{20,90,130}
\definecolor{promptNote}{RGB}{110,110,110}
\definecolor{promptKey}{RGB}{150,40,90}

\providecommand{\phead}[1]{\textcolor{promptHead}{\ttfamily\bfseries #1}}

\lstdefinestyle{prompt}{
	basicstyle=\ttfamily\scriptsize,
	extendedchars=true,
	aboveskip=0pt,
	belowskip=0pt,
	framexleftmargin=2pt,
	framexrightmargin=2pt,
	frame=single,
	rulecolor=\color{promptFrame},
	backgroundcolor=\color{promptBg},
	breaklines=true,
	breakatwhitespace=true,
	breakindent=0pt,
	postbreak=\mbox{\textcolor{promptNote}{$\hookrightarrow$}\space},
	showstringspaces=false,
	columns=fullflexible,
	keepspaces=true,
	tabsize=2,
	escapeinside={(*}{*)},
	literate=
	{ä}{{\"a}}1
	{ö}{{\"o}}1
	{ü}{{\"u}}1
	{Ä}{{\"A}}1
	{Ö}{{\"O}}1
	{Ü}{{\"U}}1
	{ß}{{\ss}}1
}

\begin{figure*}[t]
\centering
\begin{minipage}[t]{0.485\textwidth}
\begin{lstlisting}[style=prompt]
(*\phead{SPECIFIC REQUIREMENTS FOR THIS CONVERSATION}*)
- Reason for visit: patient sees a doctor about an STI
  for the first time
- Conversation style: patient is hesitant at first, [...]
- Patient profile: patient: man who has sex with men (MSM)
- Health literacy: patient has little [...]
- Conversation structure: conversation follows chronological [...]
- Setting: Checkpoint (queer health counselling)
[... atmosphere and one probabilistic flag ...]
- Patient speaks with a Berlin accent

(*\phead{PERSONS}*)
- Doctor: Lehmann
    Description: Calm, professional and efficient. Appears experienced, matter-of-fact and non-judgmental. [...] Encourages free narration.
- Patient: Kim
    Description: Is unsure at first, but opens up more and more in the course of the conversation.
\end{lstlisting}
\end{minipage}\hfill
\begin{minipage}[t]{0.485\textwidth}
\begin{lstlisting}[style=prompt]
(*\phead{GROUND-TRUTH REQUIREMENTS}*)
- All information that is NOT `not_discussed` must appear
  in the conversation
- Fields with `not_discussed` must NOT be mentioned
  (this keeps the conversation naturally short)
- Information may be conveyed explicitly or implicitly [...]
[... 4 further requirements ...]

(*\phead{TOPIC ORDER}*) (guideline, not strict)
Cover the following topics roughly in this order,
but fit them naturally into the flow of the conversation:

1. Introduction, reason for the visit, confidentiality
2. When was the last contact, how many partners recently
3. Type of practices (vaginal, anal, oral), details
4. Age, sex, anatomy, hormone therapy
[... topics 5 to 8 ...]
9. Which tests, how are results communicated
\end{lstlisting}
\end{minipage}
\caption{Excerpts from the dialogue prompt of a conversation (translated from the original German). The left column is sampled per conversation through the variation configuration and the personas. On the right, the ground-truth requirements are identical in every prompt, while the topic plan below them is derived from the sampled record. Here, nine of the fifteen declared categories are present.}
\label{fig:prompt-mechanisms}
\end{figure*}

%% file: figures/diversity-figures.tex
\begin{figure}[t]
  \centering
  \begin{minipage}[t]{0.48\textwidth}
    \centering
    \input{figures/variance_curves.pgf}
    \caption{Cumulative variance explained by the leading principal components of the conversation embeddings; a steeper curve means the variance sits in fewer directions. StiCloned and StiSimple are ours, PriMock57 and MedDialog are human. All corpora are subsampled to $n=51$ (averaged over 200 draws; size of the smallest one).}
    \label{fig:variance-curves}
  \end{minipage}\hfill
  \begin{minipage}[t]{0.48\textwidth}
    \centering
    \input{figures/tsne_overlay.pgf}
    \caption{Two-dimensional t-SNE projection of the dialogue embeddings. The earlier StiSimple dataset forms a separate group, StiCloned spreads into a diffuse cloud, and StiReal lies inside it. t-SNE does not preserve global distances, so the plot is an illustration only.}
    \label{fig:tsne-overlay}
  \end{minipage}
\end{figure}

%% file: figures/variance_curves.pgf
\begingroup%
\makeatletter%
\begin{pgfpicture}%
\pgfpathrectangle{\pgfpointorigin}{\pgfqpoint{2.855556in}{2.375556in}}%
\pgfusepath{use as bounding box, clip}%
\begin{pgfscope}%
\pgfsetbuttcap%
\pgfsetmiterjoin%
\definecolor{currentfill}{rgb}{1.000000,1.000000,1.000000}%
\pgfsetfillcolor{currentfill}%
\pgfsetlinewidth{0.000000pt}%
\definecolor{currentstroke}{rgb}{1.000000,1.000000,1.000000}%
\pgfsetstrokecolor{currentstroke}%
\pgfsetdash{}{0pt}%
\pgfpathmoveto{\pgfqpoint{0.000000in}{0.000000in}}%
\pgfpathlineto{\pgfqpoint{2.855556in}{0.000000in}}%
\pgfpathlineto{\pgfqpoint{2.855556in}{2.375556in}}%
\pgfpathlineto{\pgfqpoint{0.000000in}{2.375556in}}%
\pgfpathlineto{\pgfqpoint{0.000000in}{0.000000in}}%
\pgfpathclose%
\pgfusepath{fill}%
\end{pgfscope}%
\begin{pgfscope}%
\pgfsetbuttcap%
\pgfsetmiterjoin%
\definecolor{currentfill}{rgb}{1.000000,1.000000,1.000000}%
\pgfsetfillcolor{currentfill}%
\pgfsetlinewidth{0.000000pt}%
\definecolor{currentstroke}{rgb}{0.000000,0.000000,0.000000}%
\pgfsetstrokecolor{currentstroke}%
\pgfsetstrokeopacity{0.000000}%
\pgfsetdash{}{0pt}%
\pgfpathmoveto{\pgfqpoint{0.391366in}{0.334074in}}%
\pgfpathlineto{\pgfqpoint{2.790193in}{0.334074in}}%
\pgfpathlineto{\pgfqpoint{2.790193in}{2.331798in}}%
\pgfpathlineto{\pgfqpoint{0.391366in}{2.331798in}}%
\pgfpathlineto{\pgfqpoint{0.391366in}{0.334074in}}%
\pgfpathclose%
\pgfusepath{fill}%
\end{pgfscope}%
\begin{pgfscope}%
\pgfsetbuttcap%
\pgfsetroundjoin%
\definecolor{currentfill}{rgb}{0.000000,0.000000,0.000000}%
\pgfsetfillcolor{currentfill}%
\pgfsetlinewidth{0.602250pt}%
\definecolor{currentstroke}{rgb}{0.000000,0.000000,0.000000}%
\pgfsetstrokecolor{currentstroke}%
\pgfsetdash{}{0pt}%
\pgfsys@defobject{currentmarker}{\pgfqpoint{0.000000in}{-0.034722in}}{\pgfqpoint{0.000000in}{0.000000in}}{%
\pgfpathmoveto{\pgfqpoint{0.000000in}{0.000000in}}%
\pgfpathlineto{\pgfqpoint{0.000000in}{-0.034722in}}%
\pgfusepath{stroke,fill}%
}%
\begin{pgfscope}%
\pgfsys@transformshift{0.831967in}{0.334074in}%
\pgfsys@useobject{currentmarker}{}%
\end{pgfscope}%
\end{pgfscope}%
\begin{pgfscope}%
\definecolor{textcolor}{rgb}{0.000000,0.000000,0.000000}%
\pgfsetstrokecolor{textcolor}%
\pgfsetfillcolor{textcolor}%
\pgftext[x=0.831967in,y=0.250741in,,top]{\color{textcolor}{\rmfamily\fontsize{7.000000}{8.400000}\selectfont\catcode`\^=\active\def^{\ifmmode\sp\else\^{}\fi}\catcode`\%=\active\def
\end{pgfscope}%
\begin{pgfscope}%
\pgfsetbuttcap%
\pgfsetroundjoin%
\definecolor{currentfill}{rgb}{0.000000,0.000000,0.000000}%
\pgfsetfillcolor{currentfill}%
\pgfsetlinewidth{0.602250pt}%
\definecolor{currentstroke}{rgb}{0.000000,0.000000,0.000000}%
\pgfsetstrokecolor{currentstroke}%
\pgfsetdash{}{0pt}%
\pgfsys@defobject{currentmarker}{\pgfqpoint{0.000000in}{-0.034722in}}{\pgfqpoint{0.000000in}{0.000000in}}{%
\pgfpathmoveto{\pgfqpoint{0.000000in}{0.000000in}}%
\pgfpathlineto{\pgfqpoint{0.000000in}{-0.034722in}}%
\pgfusepath{stroke,fill}%
}%
\begin{pgfscope}%
\pgfsys@transformshift{1.321524in}{0.334074in}%
\pgfsys@useobject{currentmarker}{}%
\end{pgfscope}%
\end{pgfscope}%
\begin{pgfscope}%
\definecolor{textcolor}{rgb}{0.000000,0.000000,0.000000}%
\pgfsetstrokecolor{textcolor}%
\pgfsetfillcolor{textcolor}%
\pgftext[x=1.321524in,y=0.250741in,,top]{\color{textcolor}{\rmfamily\fontsize{7.000000}{8.400000}\selectfont\catcode`\^=\active\def^{\ifmmode\sp\else\^{}\fi}\catcode`\%=\active\def
\end{pgfscope}%
\begin{pgfscope}%
\pgfsetbuttcap%
\pgfsetroundjoin%
\definecolor{currentfill}{rgb}{0.000000,0.000000,0.000000}%
\pgfsetfillcolor{currentfill}%
\pgfsetlinewidth{0.602250pt}%
\definecolor{currentstroke}{rgb}{0.000000,0.000000,0.000000}%
\pgfsetstrokecolor{currentstroke}%
\pgfsetdash{}{0pt}%
\pgfsys@defobject{currentmarker}{\pgfqpoint{0.000000in}{-0.034722in}}{\pgfqpoint{0.000000in}{0.000000in}}{%
\pgfpathmoveto{\pgfqpoint{0.000000in}{0.000000in}}%
\pgfpathlineto{\pgfqpoint{0.000000in}{-0.034722in}}%
\pgfusepath{stroke,fill}%
}%
\begin{pgfscope}%
\pgfsys@transformshift{1.811080in}{0.334074in}%
\pgfsys@useobject{currentmarker}{}%
\end{pgfscope}%
\end{pgfscope}%
\begin{pgfscope}%
\definecolor{textcolor}{rgb}{0.000000,0.000000,0.000000}%
\pgfsetstrokecolor{textcolor}%
\pgfsetfillcolor{textcolor}%
\pgftext[x=1.811080in,y=0.250741in,,top]{\color{textcolor}{\rmfamily\fontsize{7.000000}{8.400000}\selectfont\catcode`\^=\active\def^{\ifmmode\sp\else\^{}\fi}\catcode`\%=\active\def
\end{pgfscope}%
\begin{pgfscope}%
\pgfsetbuttcap%
\pgfsetroundjoin%
\definecolor{currentfill}{rgb}{0.000000,0.000000,0.000000}%
\pgfsetfillcolor{currentfill}%
\pgfsetlinewidth{0.602250pt}%
\definecolor{currentstroke}{rgb}{0.000000,0.000000,0.000000}%
\pgfsetstrokecolor{currentstroke}%
\pgfsetdash{}{0pt}%
\pgfsys@defobject{currentmarker}{\pgfqpoint{0.000000in}{-0.034722in}}{\pgfqpoint{0.000000in}{0.000000in}}{%
\pgfpathmoveto{\pgfqpoint{0.000000in}{0.000000in}}%
\pgfpathlineto{\pgfqpoint{0.000000in}{-0.034722in}}%
\pgfusepath{stroke,fill}%
}%
\begin{pgfscope}%
\pgfsys@transformshift{2.300636in}{0.334074in}%
\pgfsys@useobject{currentmarker}{}%
\end{pgfscope}%
\end{pgfscope}%
\begin{pgfscope}%
\definecolor{textcolor}{rgb}{0.000000,0.000000,0.000000}%
\pgfsetstrokecolor{textcolor}%
\pgfsetfillcolor{textcolor}%
\pgftext[x=2.300636in,y=0.250741in,,top]{\color{textcolor}{\rmfamily\fontsize{7.000000}{8.400000}\selectfont\catcode`\^=\active\def^{\ifmmode\sp\else\^{}\fi}\catcode`\%=\active\def
\end{pgfscope}%
\begin{pgfscope}%
\pgfsetbuttcap%
\pgfsetroundjoin%
\definecolor{currentfill}{rgb}{0.000000,0.000000,0.000000}%
\pgfsetfillcolor{currentfill}%
\pgfsetlinewidth{0.602250pt}%
\definecolor{currentstroke}{rgb}{0.000000,0.000000,0.000000}%
\pgfsetstrokecolor{currentstroke}%
\pgfsetdash{}{0pt}%
\pgfsys@defobject{currentmarker}{\pgfqpoint{0.000000in}{-0.034722in}}{\pgfqpoint{0.000000in}{0.000000in}}{%
\pgfpathmoveto{\pgfqpoint{0.000000in}{0.000000in}}%
\pgfpathlineto{\pgfqpoint{0.000000in}{-0.034722in}}%
\pgfusepath{stroke,fill}%
}%
\begin{pgfscope}%
\pgfsys@transformshift{2.790193in}{0.334074in}%
\pgfsys@useobject{currentmarker}{}%
\end{pgfscope}%
\end{pgfscope}%
\begin{pgfscope}%
\definecolor{textcolor}{rgb}{0.000000,0.000000,0.000000}%
\pgfsetstrokecolor{textcolor}%
\pgfsetfillcolor{textcolor}%
\pgftext[x=2.790193in,y=0.250741in,,top]{\color{textcolor}{\rmfamily\fontsize{7.000000}{8.400000}\selectfont\catcode`\^=\active\def^{\ifmmode\sp\else\^{}\fi}\catcode`\%=\active\def
\end{pgfscope}%
\begin{pgfscope}%
\definecolor{textcolor}{rgb}{0.000000,0.000000,0.000000}%
\pgfsetstrokecolor{textcolor}%
\pgfsetfillcolor{textcolor}%
\pgftext[x=1.590780in,y=0.108766in,,top]{\color{textcolor}{\rmfamily\fontsize{8.000000}{9.600000}\selectfont\catcode`\^=\active\def^{\ifmmode\sp\else\^{}\fi}\catcode`\%=\active\def
\end{pgfscope}%
\begin{pgfscope}%
\pgfsetbuttcap%
\pgfsetroundjoin%
\definecolor{currentfill}{rgb}{0.000000,0.000000,0.000000}%
\pgfsetfillcolor{currentfill}%
\pgfsetlinewidth{0.602250pt}%
\definecolor{currentstroke}{rgb}{0.000000,0.000000,0.000000}%
\pgfsetstrokecolor{currentstroke}%
\pgfsetdash{}{0pt}%
\pgfsys@defobject{currentmarker}{\pgfqpoint{-0.034722in}{0.000000in}}{\pgfqpoint{-0.000000in}{0.000000in}}{%
\pgfpathmoveto{\pgfqpoint{-0.000000in}{0.000000in}}%
\pgfpathlineto{\pgfqpoint{-0.034722in}{0.000000in}}%
\pgfusepath{stroke,fill}%
}%
\begin{pgfscope}%
\pgfsys@transformshift{0.391366in}{0.334074in}%
\pgfsys@useobject{currentmarker}{}%
\end{pgfscope}%
\end{pgfscope}%
\begin{pgfscope}%
\definecolor{textcolor}{rgb}{0.000000,0.000000,0.000000}%
\pgfsetstrokecolor{textcolor}%
\pgfsetfillcolor{textcolor}%
\pgftext[x=0.164321in, y=0.300316in, left, base]{\color{textcolor}{\rmfamily\fontsize{7.000000}{8.400000}\selectfont\catcode`\^=\active\def^{\ifmmode\sp\else\^{}\fi}\catcode`\%=\active\def
\end{pgfscope}%
\begin{pgfscope}%
\pgfsetbuttcap%
\pgfsetroundjoin%
\definecolor{currentfill}{rgb}{0.000000,0.000000,0.000000}%
\pgfsetfillcolor{currentfill}%
\pgfsetlinewidth{0.602250pt}%
\definecolor{currentstroke}{rgb}{0.000000,0.000000,0.000000}%
\pgfsetstrokecolor{currentstroke}%
\pgfsetdash{}{0pt}%
\pgfsys@defobject{currentmarker}{\pgfqpoint{-0.034722in}{0.000000in}}{\pgfqpoint{-0.000000in}{0.000000in}}{%
\pgfpathmoveto{\pgfqpoint{-0.000000in}{0.000000in}}%
\pgfpathlineto{\pgfqpoint{-0.034722in}{0.000000in}}%
\pgfusepath{stroke,fill}%
}%
\begin{pgfscope}%
\pgfsys@transformshift{0.391366in}{0.619463in}%
\pgfsys@useobject{currentmarker}{}%
\end{pgfscope}%
\end{pgfscope}%
\begin{pgfscope}%
\definecolor{textcolor}{rgb}{0.000000,0.000000,0.000000}%
\pgfsetstrokecolor{textcolor}%
\pgfsetfillcolor{textcolor}%
\pgftext[x=0.164321in, y=0.585705in, left, base]{\color{textcolor}{\rmfamily\fontsize{7.000000}{8.400000}\selectfont\catcode`\^=\active\def^{\ifmmode\sp\else\^{}\fi}\catcode`\%=\active\def
\end{pgfscope}%
\begin{pgfscope}%
\pgfsetbuttcap%
\pgfsetroundjoin%
\definecolor{currentfill}{rgb}{0.000000,0.000000,0.000000}%
\pgfsetfillcolor{currentfill}%
\pgfsetlinewidth{0.602250pt}%
\definecolor{currentstroke}{rgb}{0.000000,0.000000,0.000000}%
\pgfsetstrokecolor{currentstroke}%
\pgfsetdash{}{0pt}%
\pgfsys@defobject{currentmarker}{\pgfqpoint{-0.034722in}{0.000000in}}{\pgfqpoint{-0.000000in}{0.000000in}}{%
\pgfpathmoveto{\pgfqpoint{-0.000000in}{0.000000in}}%
\pgfpathlineto{\pgfqpoint{-0.034722in}{0.000000in}}%
\pgfusepath{stroke,fill}%
}%
\begin{pgfscope}%
\pgfsys@transformshift{0.391366in}{0.904852in}%
\pgfsys@useobject{currentmarker}{}%
\end{pgfscope}%
\end{pgfscope}%
\begin{pgfscope}%
\definecolor{textcolor}{rgb}{0.000000,0.000000,0.000000}%
\pgfsetstrokecolor{textcolor}%
\pgfsetfillcolor{textcolor}%
\pgftext[x=0.164321in, y=0.871095in, left, base]{\color{textcolor}{\rmfamily\fontsize{7.000000}{8.400000}\selectfont\catcode`\^=\active\def^{\ifmmode\sp\else\^{}\fi}\catcode`\%=\active\def
\end{pgfscope}%
\begin{pgfscope}%
\pgfsetbuttcap%
\pgfsetroundjoin%
\definecolor{currentfill}{rgb}{0.000000,0.000000,0.000000}%
\pgfsetfillcolor{currentfill}%
\pgfsetlinewidth{0.602250pt}%
\definecolor{currentstroke}{rgb}{0.000000,0.000000,0.000000}%
\pgfsetstrokecolor{currentstroke}%
\pgfsetdash{}{0pt}%
\pgfsys@defobject{currentmarker}{\pgfqpoint{-0.034722in}{0.000000in}}{\pgfqpoint{-0.000000in}{0.000000in}}{%
\pgfpathmoveto{\pgfqpoint{-0.000000in}{0.000000in}}%
\pgfpathlineto{\pgfqpoint{-0.034722in}{0.000000in}}%
\pgfusepath{stroke,fill}%
}%
\begin{pgfscope}%
\pgfsys@transformshift{0.391366in}{1.190241in}%
\pgfsys@useobject{currentmarker}{}%
\end{pgfscope}%
\end{pgfscope}%
\begin{pgfscope}%
\definecolor{textcolor}{rgb}{0.000000,0.000000,0.000000}%
\pgfsetstrokecolor{textcolor}%
\pgfsetfillcolor{textcolor}%
\pgftext[x=0.164321in, y=1.156484in, left, base]{\color{textcolor}{\rmfamily\fontsize{7.000000}{8.400000}\selectfont\catcode`\^=\active\def^{\ifmmode\sp\else\^{}\fi}\catcode`\%=\active\def
\end{pgfscope}%
\begin{pgfscope}%
\pgfsetbuttcap%
\pgfsetroundjoin%
\definecolor{currentfill}{rgb}{0.000000,0.000000,0.000000}%
\pgfsetfillcolor{currentfill}%
\pgfsetlinewidth{0.602250pt}%
\definecolor{currentstroke}{rgb}{0.000000,0.000000,0.000000}%
\pgfsetstrokecolor{currentstroke}%
\pgfsetdash{}{0pt}%
\pgfsys@defobject{currentmarker}{\pgfqpoint{-0.034722in}{0.000000in}}{\pgfqpoint{-0.000000in}{0.000000in}}{%
\pgfpathmoveto{\pgfqpoint{-0.000000in}{0.000000in}}%
\pgfpathlineto{\pgfqpoint{-0.034722in}{0.000000in}}%
\pgfusepath{stroke,fill}%
}%
\begin{pgfscope}%
\pgfsys@transformshift{0.391366in}{1.475630in}%
\pgfsys@useobject{currentmarker}{}%
\end{pgfscope}%
\end{pgfscope}%
\begin{pgfscope}%
\definecolor{textcolor}{rgb}{0.000000,0.000000,0.000000}%
\pgfsetstrokecolor{textcolor}%
\pgfsetfillcolor{textcolor}%
\pgftext[x=0.164321in, y=1.441873in, left, base]{\color{textcolor}{\rmfamily\fontsize{7.000000}{8.400000}\selectfont\catcode`\^=\active\def^{\ifmmode\sp\else\^{}\fi}\catcode`\%=\active\def
\end{pgfscope}%
\begin{pgfscope}%
\pgfsetbuttcap%
\pgfsetroundjoin%
\definecolor{currentfill}{rgb}{0.000000,0.000000,0.000000}%
\pgfsetfillcolor{currentfill}%
\pgfsetlinewidth{0.602250pt}%
\definecolor{currentstroke}{rgb}{0.000000,0.000000,0.000000}%
\pgfsetstrokecolor{currentstroke}%
\pgfsetdash{}{0pt}%
\pgfsys@defobject{currentmarker}{\pgfqpoint{-0.034722in}{0.000000in}}{\pgfqpoint{-0.000000in}{0.000000in}}{%
\pgfpathmoveto{\pgfqpoint{-0.000000in}{0.000000in}}%
\pgfpathlineto{\pgfqpoint{-0.034722in}{0.000000in}}%
\pgfusepath{stroke,fill}%
}%
\begin{pgfscope}%
\pgfsys@transformshift{0.391366in}{1.761020in}%
\pgfsys@useobject{currentmarker}{}%
\end{pgfscope}%
\end{pgfscope}%
\begin{pgfscope}%
\definecolor{textcolor}{rgb}{0.000000,0.000000,0.000000}%
\pgfsetstrokecolor{textcolor}%
\pgfsetfillcolor{textcolor}%
\pgftext[x=0.164321in, y=1.727262in, left, base]{\color{textcolor}{\rmfamily\fontsize{7.000000}{8.400000}\selectfont\catcode`\^=\active\def^{\ifmmode\sp\else\^{}\fi}\catcode`\%=\active\def
\end{pgfscope}%
\begin{pgfscope}%
\pgfsetbuttcap%
\pgfsetroundjoin%
\definecolor{currentfill}{rgb}{0.000000,0.000000,0.000000}%
\pgfsetfillcolor{currentfill}%
\pgfsetlinewidth{0.602250pt}%
\definecolor{currentstroke}{rgb}{0.000000,0.000000,0.000000}%
\pgfsetstrokecolor{currentstroke}%
\pgfsetdash{}{0pt}%
\pgfsys@defobject{currentmarker}{\pgfqpoint{-0.034722in}{0.000000in}}{\pgfqpoint{-0.000000in}{0.000000in}}{%
\pgfpathmoveto{\pgfqpoint{-0.000000in}{0.000000in}}%
\pgfpathlineto{\pgfqpoint{-0.034722in}{0.000000in}}%
\pgfusepath{stroke,fill}%
}%
\begin{pgfscope}%
\pgfsys@transformshift{0.391366in}{2.046409in}%
\pgfsys@useobject{currentmarker}{}%
\end{pgfscope}%
\end{pgfscope}%
\begin{pgfscope}%
\definecolor{textcolor}{rgb}{0.000000,0.000000,0.000000}%
\pgfsetstrokecolor{textcolor}%
\pgfsetfillcolor{textcolor}%
\pgftext[x=0.164321in, y=2.012651in, left, base]{\color{textcolor}{\rmfamily\fontsize{7.000000}{8.400000}\selectfont\catcode`\^=\active\def^{\ifmmode\sp\else\^{}\fi}\catcode`\%=\active\def
\end{pgfscope}%
\begin{pgfscope}%
\pgfsetbuttcap%
\pgfsetroundjoin%
\definecolor{currentfill}{rgb}{0.000000,0.000000,0.000000}%
\pgfsetfillcolor{currentfill}%
\pgfsetlinewidth{0.602250pt}%
\definecolor{currentstroke}{rgb}{0.000000,0.000000,0.000000}%
\pgfsetstrokecolor{currentstroke}%
\pgfsetdash{}{0pt}%
\pgfsys@defobject{currentmarker}{\pgfqpoint{-0.034722in}{0.000000in}}{\pgfqpoint{-0.000000in}{0.000000in}}{%
\pgfpathmoveto{\pgfqpoint{-0.000000in}{0.000000in}}%
\pgfpathlineto{\pgfqpoint{-0.034722in}{0.000000in}}%
\pgfusepath{stroke,fill}%
}%
\begin{pgfscope}%
\pgfsys@transformshift{0.391366in}{2.331798in}%
\pgfsys@useobject{currentmarker}{}%
\end{pgfscope}%
\end{pgfscope}%
\begin{pgfscope}%
\definecolor{textcolor}{rgb}{0.000000,0.000000,0.000000}%
\pgfsetstrokecolor{textcolor}%
\pgfsetfillcolor{textcolor}%
\pgftext[x=0.164321in, y=2.298040in, left, base]{\color{textcolor}{\rmfamily\fontsize{7.000000}{8.400000}\selectfont\catcode`\^=\active\def^{\ifmmode\sp\else\^{}\fi}\catcode`\%=\active\def
\end{pgfscope}%
\begin{pgfscope}%
\definecolor{textcolor}{rgb}{0.000000,0.000000,0.000000}%
\pgfsetstrokecolor{textcolor}%
\pgfsetfillcolor{textcolor}%
\pgftext[x=0.108766in,y=1.332936in,,bottom,rotate=90.000000]{\color{textcolor}{\rmfamily\fontsize{8.000000}{9.600000}\selectfont\catcode`\^=\active\def^{\ifmmode\sp\else\^{}\fi}\catcode`\%=\active\def
\end{pgfscope}%
\begin{pgfscope}%
\pgfpathrectangle{\pgfqpoint{0.391366in}{0.334074in}}{\pgfqpoint{2.398826in}{1.997724in}}%
\pgfusepath{clip}%
\pgfsetrectcap%
\pgfsetroundjoin%
\pgfsetlinewidth{1.104125pt}%
\definecolor{currentstroke}{rgb}{0.231373,0.431373,0.647059}%
\pgfsetstrokecolor{currentstroke}%
\pgfsetdash{}{0pt}%
\pgfpathmoveto{\pgfqpoint{0.515859in}{0.324074in}}%
\pgfpathlineto{\pgfqpoint{0.538233in}{0.397398in}}%
\pgfpathlineto{\pgfqpoint{0.587189in}{0.534755in}}%
\pgfpathlineto{\pgfqpoint{0.636145in}{0.654993in}}%
\pgfpathlineto{\pgfqpoint{0.685100in}{0.761977in}}%
\pgfpathlineto{\pgfqpoint{0.734056in}{0.857571in}}%
\pgfpathlineto{\pgfqpoint{0.783011in}{0.944491in}}%
\pgfpathlineto{\pgfqpoint{0.831967in}{1.024569in}}%
\pgfpathlineto{\pgfqpoint{0.880923in}{1.098573in}}%
\pgfpathlineto{\pgfqpoint{0.929878in}{1.167393in}}%
\pgfpathlineto{\pgfqpoint{0.978834in}{1.232070in}}%
\pgfpathlineto{\pgfqpoint{1.027790in}{1.292890in}}%
\pgfpathlineto{\pgfqpoint{1.076745in}{1.350336in}}%
\pgfpathlineto{\pgfqpoint{1.125701in}{1.404782in}}%
\pgfpathlineto{\pgfqpoint{1.174657in}{1.456373in}}%
\pgfpathlineto{\pgfqpoint{1.223612in}{1.505431in}}%
\pgfpathlineto{\pgfqpoint{1.272568in}{1.552195in}}%
\pgfpathlineto{\pgfqpoint{1.321524in}{1.596843in}}%
\pgfpathlineto{\pgfqpoint{1.370479in}{1.639407in}}%
\pgfpathlineto{\pgfqpoint{1.419435in}{1.680199in}}%
\pgfpathlineto{\pgfqpoint{1.468390in}{1.719151in}}%
\pgfpathlineto{\pgfqpoint{1.517346in}{1.756451in}}%
\pgfpathlineto{\pgfqpoint{1.566302in}{1.792300in}}%
\pgfpathlineto{\pgfqpoint{1.615257in}{1.826707in}}%
\pgfpathlineto{\pgfqpoint{1.664213in}{1.859707in}}%
\pgfpathlineto{\pgfqpoint{1.713169in}{1.891366in}}%
\pgfpathlineto{\pgfqpoint{1.762124in}{1.921671in}}%
\pgfpathlineto{\pgfqpoint{1.811080in}{1.950816in}}%
\pgfpathlineto{\pgfqpoint{1.860036in}{1.978842in}}%
\pgfpathlineto{\pgfqpoint{1.908991in}{2.005743in}}%
\pgfpathlineto{\pgfqpoint{1.957947in}{2.031600in}}%
\pgfpathlineto{\pgfqpoint{2.006902in}{2.056377in}}%
\pgfpathlineto{\pgfqpoint{2.055858in}{2.080118in}}%
\pgfpathlineto{\pgfqpoint{2.104814in}{2.102924in}}%
\pgfpathlineto{\pgfqpoint{2.153769in}{2.124758in}}%
\pgfpathlineto{\pgfqpoint{2.202725in}{2.145678in}}%
\pgfpathlineto{\pgfqpoint{2.251681in}{2.165659in}}%
\pgfpathlineto{\pgfqpoint{2.300636in}{2.184776in}}%
\pgfpathlineto{\pgfqpoint{2.349592in}{2.203077in}}%
\pgfpathlineto{\pgfqpoint{2.398548in}{2.220568in}}%
\pgfpathlineto{\pgfqpoint{2.447503in}{2.237263in}}%
\pgfpathlineto{\pgfqpoint{2.496459in}{2.253207in}}%
\pgfpathlineto{\pgfqpoint{2.545414in}{2.268370in}}%
\pgfpathlineto{\pgfqpoint{2.594370in}{2.282705in}}%
\pgfpathlineto{\pgfqpoint{2.643326in}{2.296249in}}%
\pgfpathlineto{\pgfqpoint{2.692281in}{2.308997in}}%
\pgfpathlineto{\pgfqpoint{2.741237in}{2.320881in}}%
\pgfpathlineto{\pgfqpoint{2.790193in}{2.331798in}}%
\pgfusepath{stroke}%
\end{pgfscope}%
\begin{pgfscope}%
\pgfpathrectangle{\pgfqpoint{0.391366in}{0.334074in}}{\pgfqpoint{2.398826in}{1.997724in}}%
\pgfusepath{clip}%
\pgfsetbuttcap%
\pgfsetroundjoin%
\pgfsetlinewidth{1.104125pt}%
\definecolor{currentstroke}{rgb}{0.768627,0.384314,0.176471}%
\pgfsetstrokecolor{currentstroke}%
\pgfsetdash{{4.070000pt}{1.760000pt}}{0.000000pt}%
\pgfpathmoveto{\pgfqpoint{0.403004in}{0.324074in}}%
\pgfpathlineto{\pgfqpoint{0.440322in}{0.666918in}}%
\pgfpathlineto{\pgfqpoint{0.489278in}{0.880285in}}%
\pgfpathlineto{\pgfqpoint{0.538233in}{1.060905in}}%
\pgfpathlineto{\pgfqpoint{0.587189in}{1.189550in}}%
\pgfpathlineto{\pgfqpoint{0.636145in}{1.276766in}}%
\pgfpathlineto{\pgfqpoint{0.685100in}{1.356061in}}%
\pgfpathlineto{\pgfqpoint{0.734056in}{1.429075in}}%
\pgfpathlineto{\pgfqpoint{0.783011in}{1.490928in}}%
\pgfpathlineto{\pgfqpoint{0.831967in}{1.550176in}}%
\pgfpathlineto{\pgfqpoint{0.880923in}{1.596603in}}%
\pgfpathlineto{\pgfqpoint{0.929878in}{1.638285in}}%
\pgfpathlineto{\pgfqpoint{0.978834in}{1.678319in}}%
\pgfpathlineto{\pgfqpoint{1.027790in}{1.713802in}}%
\pgfpathlineto{\pgfqpoint{1.076745in}{1.746476in}}%
\pgfpathlineto{\pgfqpoint{1.125701in}{1.778553in}}%
\pgfpathlineto{\pgfqpoint{1.174657in}{1.809425in}}%
\pgfpathlineto{\pgfqpoint{1.223612in}{1.838250in}}%
\pgfpathlineto{\pgfqpoint{1.272568in}{1.864768in}}%
\pgfpathlineto{\pgfqpoint{1.321524in}{1.890810in}}%
\pgfpathlineto{\pgfqpoint{1.370479in}{1.915516in}}%
\pgfpathlineto{\pgfqpoint{1.419435in}{1.938901in}}%
\pgfpathlineto{\pgfqpoint{1.468390in}{1.961833in}}%
\pgfpathlineto{\pgfqpoint{1.517346in}{1.984143in}}%
\pgfpathlineto{\pgfqpoint{1.566302in}{2.005039in}}%
\pgfpathlineto{\pgfqpoint{1.615257in}{2.025675in}}%
\pgfpathlineto{\pgfqpoint{1.664213in}{2.045906in}}%
\pgfpathlineto{\pgfqpoint{1.713169in}{2.064770in}}%
\pgfpathlineto{\pgfqpoint{1.762124in}{2.083237in}}%
\pgfpathlineto{\pgfqpoint{1.811080in}{2.100542in}}%
\pgfpathlineto{\pgfqpoint{1.860036in}{2.116906in}}%
\pgfpathlineto{\pgfqpoint{1.908991in}{2.132734in}}%
\pgfpathlineto{\pgfqpoint{1.957947in}{2.147859in}}%
\pgfpathlineto{\pgfqpoint{2.006902in}{2.162642in}}%
\pgfpathlineto{\pgfqpoint{2.055858in}{2.177068in}}%
\pgfpathlineto{\pgfqpoint{2.104814in}{2.190616in}}%
\pgfpathlineto{\pgfqpoint{2.153769in}{2.203813in}}%
\pgfpathlineto{\pgfqpoint{2.202725in}{2.216086in}}%
\pgfpathlineto{\pgfqpoint{2.251681in}{2.227622in}}%
\pgfpathlineto{\pgfqpoint{2.300636in}{2.238910in}}%
\pgfpathlineto{\pgfqpoint{2.349592in}{2.250080in}}%
\pgfpathlineto{\pgfqpoint{2.398548in}{2.260715in}}%
\pgfpathlineto{\pgfqpoint{2.447503in}{2.271219in}}%
\pgfpathlineto{\pgfqpoint{2.496459in}{2.281516in}}%
\pgfpathlineto{\pgfqpoint{2.545414in}{2.291437in}}%
\pgfpathlineto{\pgfqpoint{2.594370in}{2.300769in}}%
\pgfpathlineto{\pgfqpoint{2.643326in}{2.309283in}}%
\pgfpathlineto{\pgfqpoint{2.692281in}{2.317417in}}%
\pgfpathlineto{\pgfqpoint{2.741237in}{2.324788in}}%
\pgfpathlineto{\pgfqpoint{2.790193in}{2.331799in}}%
\pgfusepath{stroke}%
\end{pgfscope}%
\begin{pgfscope}%
\pgfpathrectangle{\pgfqpoint{0.391366in}{0.334074in}}{\pgfqpoint{2.398826in}{1.997724in}}%
\pgfusepath{clip}%
\pgfsetbuttcap%
\pgfsetroundjoin%
\pgfsetlinewidth{1.104125pt}%
\definecolor{currentstroke}{rgb}{0.105882,0.498039,0.388235}%
\pgfsetstrokecolor{currentstroke}%
\pgfsetdash{{7.040000pt}{1.760000pt}{1.100000pt}{1.760000pt}}{0.000000pt}%
\pgfpathmoveto{\pgfqpoint{0.478480in}{0.324074in}}%
\pgfpathlineto{\pgfqpoint{0.489278in}{0.377581in}}%
\pgfpathlineto{\pgfqpoint{0.538233in}{0.577461in}}%
\pgfpathlineto{\pgfqpoint{0.587189in}{0.740832in}}%
\pgfpathlineto{\pgfqpoint{0.636145in}{0.871229in}}%
\pgfpathlineto{\pgfqpoint{0.685100in}{0.991685in}}%
\pgfpathlineto{\pgfqpoint{0.734056in}{1.097946in}}%
\pgfpathlineto{\pgfqpoint{0.783011in}{1.195583in}}%
\pgfpathlineto{\pgfqpoint{0.831967in}{1.280821in}}%
\pgfpathlineto{\pgfqpoint{0.880923in}{1.358201in}}%
\pgfpathlineto{\pgfqpoint{0.929878in}{1.425862in}}%
\pgfpathlineto{\pgfqpoint{0.978834in}{1.488471in}}%
\pgfpathlineto{\pgfqpoint{1.027790in}{1.545699in}}%
\pgfpathlineto{\pgfqpoint{1.076745in}{1.598955in}}%
\pgfpathlineto{\pgfqpoint{1.125701in}{1.648470in}}%
\pgfpathlineto{\pgfqpoint{1.174657in}{1.695119in}}%
\pgfpathlineto{\pgfqpoint{1.223612in}{1.738553in}}%
\pgfpathlineto{\pgfqpoint{1.272568in}{1.778791in}}%
\pgfpathlineto{\pgfqpoint{1.321524in}{1.816051in}}%
\pgfpathlineto{\pgfqpoint{1.370479in}{1.850367in}}%
\pgfpathlineto{\pgfqpoint{1.419435in}{1.882350in}}%
\pgfpathlineto{\pgfqpoint{1.468390in}{1.912235in}}%
\pgfpathlineto{\pgfqpoint{1.517346in}{1.940207in}}%
\pgfpathlineto{\pgfqpoint{1.566302in}{1.966368in}}%
\pgfpathlineto{\pgfqpoint{1.615257in}{1.991068in}}%
\pgfpathlineto{\pgfqpoint{1.664213in}{2.014158in}}%
\pgfpathlineto{\pgfqpoint{1.713169in}{2.036078in}}%
\pgfpathlineto{\pgfqpoint{1.762124in}{2.056839in}}%
\pgfpathlineto{\pgfqpoint{1.811080in}{2.076581in}}%
\pgfpathlineto{\pgfqpoint{1.860036in}{2.095476in}}%
\pgfpathlineto{\pgfqpoint{1.908991in}{2.113552in}}%
\pgfpathlineto{\pgfqpoint{1.957947in}{2.130769in}}%
\pgfpathlineto{\pgfqpoint{2.006902in}{2.147196in}}%
\pgfpathlineto{\pgfqpoint{2.055858in}{2.162916in}}%
\pgfpathlineto{\pgfqpoint{2.104814in}{2.177986in}}%
\pgfpathlineto{\pgfqpoint{2.153769in}{2.192360in}}%
\pgfpathlineto{\pgfqpoint{2.202725in}{2.206040in}}%
\pgfpathlineto{\pgfqpoint{2.251681in}{2.219128in}}%
\pgfpathlineto{\pgfqpoint{2.300636in}{2.231673in}}%
\pgfpathlineto{\pgfqpoint{2.349592in}{2.243730in}}%
\pgfpathlineto{\pgfqpoint{2.398548in}{2.255329in}}%
\pgfpathlineto{\pgfqpoint{2.447503in}{2.266461in}}%
\pgfpathlineto{\pgfqpoint{2.496459in}{2.277132in}}%
\pgfpathlineto{\pgfqpoint{2.545414in}{2.287400in}}%
\pgfpathlineto{\pgfqpoint{2.594370in}{2.297221in}}%
\pgfpathlineto{\pgfqpoint{2.643326in}{2.306577in}}%
\pgfpathlineto{\pgfqpoint{2.692281in}{2.315434in}}%
\pgfpathlineto{\pgfqpoint{2.741237in}{2.323850in}}%
\pgfpathlineto{\pgfqpoint{2.790193in}{2.331798in}}%
\pgfusepath{stroke}%
\end{pgfscope}%
\begin{pgfscope}%
\pgfpathrectangle{\pgfqpoint{0.391366in}{0.334074in}}{\pgfqpoint{2.398826in}{1.997724in}}%
\pgfusepath{clip}%
\pgfsetbuttcap%
\pgfsetroundjoin%
\pgfsetlinewidth{1.104125pt}%
\definecolor{currentstroke}{rgb}{0.541176,0.227451,0.360784}%
\pgfsetstrokecolor{currentstroke}%
\pgfsetdash{{1.100000pt}{1.815000pt}}{0.000000pt}%
\pgfpathmoveto{\pgfqpoint{0.529770in}{0.324074in}}%
\pgfpathlineto{\pgfqpoint{0.538233in}{0.350958in}}%
\pgfpathlineto{\pgfqpoint{0.587189in}{0.485190in}}%
\pgfpathlineto{\pgfqpoint{0.636145in}{0.604956in}}%
\pgfpathlineto{\pgfqpoint{0.685100in}{0.713810in}}%
\pgfpathlineto{\pgfqpoint{0.734056in}{0.812777in}}%
\pgfpathlineto{\pgfqpoint{0.783011in}{0.904203in}}%
\pgfpathlineto{\pgfqpoint{0.831967in}{0.989033in}}%
\pgfpathlineto{\pgfqpoint{0.880923in}{1.067943in}}%
\pgfpathlineto{\pgfqpoint{0.929878in}{1.141939in}}%
\pgfpathlineto{\pgfqpoint{0.978834in}{1.211178in}}%
\pgfpathlineto{\pgfqpoint{1.027790in}{1.276162in}}%
\pgfpathlineto{\pgfqpoint{1.076745in}{1.337587in}}%
\pgfpathlineto{\pgfqpoint{1.125701in}{1.395572in}}%
\pgfpathlineto{\pgfqpoint{1.174657in}{1.450430in}}%
\pgfpathlineto{\pgfqpoint{1.223612in}{1.502555in}}%
\pgfpathlineto{\pgfqpoint{1.272568in}{1.552020in}}%
\pgfpathlineto{\pgfqpoint{1.321524in}{1.599098in}}%
\pgfpathlineto{\pgfqpoint{1.370479in}{1.643805in}}%
\pgfpathlineto{\pgfqpoint{1.419435in}{1.686412in}}%
\pgfpathlineto{\pgfqpoint{1.468390in}{1.727022in}}%
\pgfpathlineto{\pgfqpoint{1.517346in}{1.765872in}}%
\pgfpathlineto{\pgfqpoint{1.566302in}{1.803067in}}%
\pgfpathlineto{\pgfqpoint{1.615257in}{1.838524in}}%
\pgfpathlineto{\pgfqpoint{1.664213in}{1.872363in}}%
\pgfpathlineto{\pgfqpoint{1.713169in}{1.904784in}}%
\pgfpathlineto{\pgfqpoint{1.762124in}{1.935772in}}%
\pgfpathlineto{\pgfqpoint{1.811080in}{1.965317in}}%
\pgfpathlineto{\pgfqpoint{1.860036in}{1.993598in}}%
\pgfpathlineto{\pgfqpoint{1.908991in}{2.020624in}}%
\pgfpathlineto{\pgfqpoint{1.957947in}{2.046477in}}%
\pgfpathlineto{\pgfqpoint{2.006902in}{2.071145in}}%
\pgfpathlineto{\pgfqpoint{2.055858in}{2.094725in}}%
\pgfpathlineto{\pgfqpoint{2.104814in}{2.117289in}}%
\pgfpathlineto{\pgfqpoint{2.153769in}{2.138852in}}%
\pgfpathlineto{\pgfqpoint{2.202725in}{2.159377in}}%
\pgfpathlineto{\pgfqpoint{2.251681in}{2.178994in}}%
\pgfpathlineto{\pgfqpoint{2.300636in}{2.197580in}}%
\pgfpathlineto{\pgfqpoint{2.349592in}{2.215308in}}%
\pgfpathlineto{\pgfqpoint{2.398548in}{2.232162in}}%
\pgfpathlineto{\pgfqpoint{2.447503in}{2.248087in}}%
\pgfpathlineto{\pgfqpoint{2.496459in}{2.263119in}}%
\pgfpathlineto{\pgfqpoint{2.545414in}{2.277241in}}%
\pgfpathlineto{\pgfqpoint{2.594370in}{2.290412in}}%
\pgfpathlineto{\pgfqpoint{2.643326in}{2.302622in}}%
\pgfpathlineto{\pgfqpoint{2.692281in}{2.313779in}}%
\pgfpathlineto{\pgfqpoint{2.741237in}{2.323702in}}%
\pgfpathlineto{\pgfqpoint{2.790193in}{2.331798in}}%
\pgfusepath{stroke}%
\end{pgfscope}%
\begin{pgfscope}%
\pgfpathrectangle{\pgfqpoint{0.391366in}{0.334074in}}{\pgfqpoint{2.398826in}{1.997724in}}%
\pgfusepath{clip}%
\pgfsetbuttcap%
\pgfsetroundjoin%
\pgfsetlinewidth{0.501875pt}%
\definecolor{currentstroke}{rgb}{0.600000,0.600000,0.600000}%
\pgfsetstrokecolor{currentstroke}%
\pgfsetdash{{0.500000pt}{1.000000pt}}{0.000000pt}%
\pgfpathmoveto{\pgfqpoint{0.391366in}{2.046409in}}%
\pgfpathlineto{\pgfqpoint{2.790193in}{2.046409in}}%
\pgfusepath{stroke}%
\end{pgfscope}%
\begin{pgfscope}%
\pgfpathrectangle{\pgfqpoint{0.391366in}{0.334074in}}{\pgfqpoint{2.398826in}{1.997724in}}%
\pgfusepath{clip}%
\pgfsetbuttcap%
\pgfsetroundjoin%
\pgfsetlinewidth{0.501875pt}%
\definecolor{currentstroke}{rgb}{0.600000,0.600000,0.600000}%
\pgfsetstrokecolor{currentstroke}%
\pgfsetdash{{0.500000pt}{1.000000pt}}{0.000000pt}%
\pgfpathmoveto{\pgfqpoint{0.391366in}{2.189103in}}%
\pgfpathlineto{\pgfqpoint{2.790193in}{2.189103in}}%
\pgfusepath{stroke}%
\end{pgfscope}%
\begin{pgfscope}%
\pgfsetrectcap%
\pgfsetmiterjoin%
\pgfsetlinewidth{0.602250pt}%
\definecolor{currentstroke}{rgb}{0.600000,0.600000,0.600000}%
\pgfsetstrokecolor{currentstroke}%
\pgfsetdash{}{0pt}%
\pgfpathmoveto{\pgfqpoint{0.391366in}{0.334074in}}%
\pgfpathlineto{\pgfqpoint{0.391366in}{2.331798in}}%
\pgfusepath{stroke}%
\end{pgfscope}%
\begin{pgfscope}%
\pgfsetrectcap%
\pgfsetmiterjoin%
\pgfsetlinewidth{0.602250pt}%
\definecolor{currentstroke}{rgb}{0.600000,0.600000,0.600000}%
\pgfsetstrokecolor{currentstroke}%
\pgfsetdash{}{0pt}%
\pgfpathmoveto{\pgfqpoint{2.790193in}{0.334074in}}%
\pgfpathlineto{\pgfqpoint{2.790193in}{2.331798in}}%
\pgfusepath{stroke}%
\end{pgfscope}%
\begin{pgfscope}%
\pgfsetrectcap%
\pgfsetmiterjoin%
\pgfsetlinewidth{0.602250pt}%
\definecolor{currentstroke}{rgb}{0.600000,0.600000,0.600000}%
\pgfsetstrokecolor{currentstroke}%
\pgfsetdash{}{0pt}%
\pgfpathmoveto{\pgfqpoint{0.391366in}{0.334074in}}%
\pgfpathlineto{\pgfqpoint{2.790193in}{0.334074in}}%
\pgfusepath{stroke}%
\end{pgfscope}%
\begin{pgfscope}%
\pgfsetrectcap%
\pgfsetmiterjoin%
\pgfsetlinewidth{0.602250pt}%
\definecolor{currentstroke}{rgb}{0.600000,0.600000,0.600000}%
\pgfsetstrokecolor{currentstroke}%
\pgfsetdash{}{0pt}%
\pgfpathmoveto{\pgfqpoint{0.391366in}{2.331798in}}%
\pgfpathlineto{\pgfqpoint{2.790193in}{2.331798in}}%
\pgfusepath{stroke}%
\end{pgfscope}%
\begin{pgfscope}%
\definecolor{textcolor}{rgb}{0.400000,0.400000,0.400000}%
\pgfsetstrokecolor{textcolor}%
\pgfsetfillcolor{textcolor}%
\pgftext[x=0.427349in,y=2.060678in,left,bottom]{\color{textcolor}{\rmfamily\fontsize{7.000000}{8.400000}\selectfont\catcode`\^=\active\def^{\ifmmode\sp\else\^{}\fi}\catcode`\%=\active\def
\end{pgfscope}%
\begin{pgfscope}%
\definecolor{textcolor}{rgb}{0.400000,0.400000,0.400000}%
\pgfsetstrokecolor{textcolor}%
\pgfsetfillcolor{textcolor}%
\pgftext[x=0.427349in,y=2.203373in,left,bottom]{\color{textcolor}{\rmfamily\fontsize{7.000000}{8.400000}\selectfont\catcode`\^=\active\def^{\ifmmode\sp\else\^{}\fi}\catcode`\%=\active\def
\end{pgfscope}%
\begin{pgfscope}%
\pgfsetbuttcap%
\pgfsetmiterjoin%
\definecolor{currentfill}{rgb}{1.000000,1.000000,1.000000}%
\pgfsetfillcolor{currentfill}%
\pgfsetfillopacity{0.900000}%
\pgfsetlinewidth{1.003750pt}%
\definecolor{currentstroke}{rgb}{0.800000,0.800000,0.800000}%
\pgfsetstrokecolor{currentstroke}%
\pgfsetstrokeopacity{0.900000}%
\pgfsetdash{}{0pt}%
\pgfpathmoveto{\pgfqpoint{1.936390in}{0.382685in}}%
\pgfpathlineto{\pgfqpoint{2.722137in}{0.382685in}}%
\pgfpathquadraticcurveto{\pgfqpoint{2.741582in}{0.382685in}}{\pgfqpoint{2.741582in}{0.402129in}}%
\pgfpathlineto{\pgfqpoint{2.741582in}{0.915246in}}%
\pgfpathquadraticcurveto{\pgfqpoint{2.741582in}{0.934691in}}{\pgfqpoint{2.722137in}{0.934691in}}%
\pgfpathlineto{\pgfqpoint{1.936390in}{0.934691in}}%
\pgfpathquadraticcurveto{\pgfqpoint{1.916945in}{0.934691in}}{\pgfqpoint{1.916945in}{0.915246in}}%
\pgfpathlineto{\pgfqpoint{1.916945in}{0.402129in}}%
\pgfpathquadraticcurveto{\pgfqpoint{1.916945in}{0.382685in}}{\pgfqpoint{1.936390in}{0.382685in}}%
\pgfpathlineto{\pgfqpoint{1.936390in}{0.382685in}}%
\pgfpathclose%
\pgfusepath{stroke,fill}%
\end{pgfscope}%
\begin{pgfscope}%
\pgfsetrectcap%
\pgfsetroundjoin%
\pgfsetlinewidth{1.104125pt}%
\definecolor{currentstroke}{rgb}{0.231373,0.431373,0.647059}%
\pgfsetstrokecolor{currentstroke}%
\pgfsetdash{}{0pt}%
\pgfpathmoveto{\pgfqpoint{1.946112in}{0.871496in}}%
\pgfpathlineto{\pgfqpoint{2.043334in}{0.871496in}}%
\pgfpathlineto{\pgfqpoint{2.140557in}{0.871496in}}%
\pgfusepath{stroke}%
\end{pgfscope}%
\begin{pgfscope}%
\definecolor{textcolor}{rgb}{0.000000,0.000000,0.000000}%
\pgfsetstrokecolor{textcolor}%
\pgfsetfillcolor{textcolor}%
\pgftext[x=2.189168in,y=0.837468in,left,base]{\color{textcolor}{\rmfamily\fontsize{7.000000}{8.400000}\selectfont\catcode`\^=\active\def^{\ifmmode\sp\else\^{}\fi}\catcode`\%=\active\def
\end{pgfscope}%
\begin{pgfscope}%
\pgfsetbuttcap%
\pgfsetroundjoin%
\pgfsetlinewidth{1.104125pt}%
\definecolor{currentstroke}{rgb}{0.768627,0.384314,0.176471}%
\pgfsetstrokecolor{currentstroke}%
\pgfsetdash{{4.070000pt}{1.760000pt}}{0.000000pt}%
\pgfpathmoveto{\pgfqpoint{1.946112in}{0.735925in}}%
\pgfpathlineto{\pgfqpoint{2.043334in}{0.735925in}}%
\pgfpathlineto{\pgfqpoint{2.140557in}{0.735925in}}%
\pgfusepath{stroke}%
\end{pgfscope}%
\begin{pgfscope}%
\definecolor{textcolor}{rgb}{0.000000,0.000000,0.000000}%
\pgfsetstrokecolor{textcolor}%
\pgfsetfillcolor{textcolor}%
\pgftext[x=2.189168in,y=0.701897in,left,base]{\color{textcolor}{\rmfamily\fontsize{7.000000}{8.400000}\selectfont\catcode`\^=\active\def^{\ifmmode\sp\else\^{}\fi}\catcode`\%=\active\def
\end{pgfscope}%
\begin{pgfscope}%
\pgfsetbuttcap%
\pgfsetroundjoin%
\pgfsetlinewidth{1.104125pt}%
\definecolor{currentstroke}{rgb}{0.105882,0.498039,0.388235}%
\pgfsetstrokecolor{currentstroke}%
\pgfsetdash{{7.040000pt}{1.760000pt}{1.100000pt}{1.760000pt}}{0.000000pt}%
\pgfpathmoveto{\pgfqpoint{1.946112in}{0.600354in}}%
\pgfpathlineto{\pgfqpoint{2.043334in}{0.600354in}}%
\pgfpathlineto{\pgfqpoint{2.140557in}{0.600354in}}%
\pgfusepath{stroke}%
\end{pgfscope}%
\begin{pgfscope}%
\definecolor{textcolor}{rgb}{0.000000,0.000000,0.000000}%
\pgfsetstrokecolor{textcolor}%
\pgfsetfillcolor{textcolor}%
\pgftext[x=2.189168in,y=0.566327in,left,base]{\color{textcolor}{\rmfamily\fontsize{7.000000}{8.400000}\selectfont\catcode`\^=\active\def^{\ifmmode\sp\else\^{}\fi}\catcode`\%=\active\def
\end{pgfscope}%
\begin{pgfscope}%
\pgfsetbuttcap%
\pgfsetroundjoin%
\pgfsetlinewidth{1.104125pt}%
\definecolor{currentstroke}{rgb}{0.541176,0.227451,0.360784}%
\pgfsetstrokecolor{currentstroke}%
\pgfsetdash{{1.100000pt}{1.815000pt}}{0.000000pt}%
\pgfpathmoveto{\pgfqpoint{1.946112in}{0.464784in}}%
\pgfpathlineto{\pgfqpoint{2.043334in}{0.464784in}}%
\pgfpathlineto{\pgfqpoint{2.140557in}{0.464784in}}%
\pgfusepath{stroke}%
\end{pgfscope}%
\begin{pgfscope}%
\definecolor{textcolor}{rgb}{0.000000,0.000000,0.000000}%
\pgfsetstrokecolor{textcolor}%
\pgfsetfillcolor{textcolor}%
\pgftext[x=2.189168in,y=0.430756in,left,base]{\color{textcolor}{\rmfamily\fontsize{7.000000}{8.400000}\selectfont\catcode`\^=\active\def^{\ifmmode\sp\else\^{}\fi}\catcode`\%=\active\def
\end{pgfscope}%
\end{pgfpicture}%
\makeatother%
\endgroup%